\documentclass[11pt,a4paper]{article}
\usepackage[hyperref]{emnlp2020}
\usepackage{times, graphicx}
\usepackage{latexsym}

\usepackage{url}
\usepackage{microtype}

\aclfinalcopy 
\def\aclpaperid{48} 

\title{Scientific Document Summarization for LaySumm$\> $'20 and LongSumm$\> $'20}

\author{Sayar Ghosh Roy, Nikhil Pinnaparaju, Risubh Jain, Manish Gupta\thanks{\ \ The author also works as a researcher at Microsoft}\ \ and Vasudeva Varma  \vspace{5pt}
\\Information Retrieval and Extraction Lab
\\International Institute of Information Technology, Hyderabad, India  \vspace{5pt}
\\ \small{\texttt{\{sayar.ghosh, nikhil.pinnaparaju\}@research.iiit.ac.in,}}
\\ \small{\texttt{risubh.jain@students.iiit.ac.in, \{manish.gupta, vv\}@iiit.ac.in}}
}
\date{}
\begin{document}
\maketitle
\begin{abstract}
Automatic text summarization has been widely studied as an important task in natural language processing. 
Traditionally, various feature engineering and machine learning based systems have been proposed for extractive as well as abstractive text summarization. Recently, deep learning based, specifically Transformer-based systems have been immensely popular.
Summarization is a cognitively challenging task -- extracting summary worthy sentences is laborious, and expressing semantics in brief when doing abstractive summarization is complicated. 
In this paper, we specifically look at the problem of summarizing scientific research papers from multiple domains. We differentiate between two types of summaries, namely, (a) LaySumm: A very short summary that captures the essence of the research paper in layman terms restricting overtly specific technical jargon and (b) LongSumm: A much longer detailed summary aimed at providing specific insights into various ideas touched upon in the paper.
While leveraging latest Transformer-based models, our systems are simple, intuitive and based on how specific paper sections contribute to human summaries of the two types described above.
Evaluations against gold standard summaries using ROUGE~\cite{rouge} metrics prove the effectiveness of our approach. On blind test corpora, our system ranks first and third for the LongSumm and LaySumm tasks respectively.
\end{abstract}

\section{Introduction}
Popularity of data science in recent years has led to a massive growth in the number of published papers online. This has generated an epochal change in the way we retrieve, analyze and consume information from these papers. Also wider interest in data science implies even lay persons (readers outside the data science community) are significantly interested in keeping up with the latest developments. The readers have access to a huge amount of such research papers on the web. For a human, understanding large documents and assimilating crucial information out of them is often a laborious and time-consuming task. Motivation to make a concise representation of huge text while retaining the core meaning of the original text has led to the development of various automated summarization systems. These systems provide users filtered, high-quality and concise content to work with at an unprecedented scale and speed. Summarization methods are mainly classified into two categories: \textit{extractive} and \textit{abstractive}. Extractive methods aim to select salient phrases, sentences or elements from the text while abstractive techniques focus on generating summaries from scratch without the constraint of reusing phrases from the original text.

Scientific papers are large, complex documents that tend to be geared towards a particular audience. This is a very small percentage of the population while majority of individuals are unable to fully comprehend the contents of long scientific documents. Even among the people who are able to understand the material, the length of such documents often spanning several pages demand a great deal of time and attention. Hence, tasks like layman summarization (LaySumm) and long-form summarization (LongSumm) are of great importance in today's world. 

Typically scientific research papers are fairly structured documents containing standard sections like abstract, introduction, background, related work, experiments, results, discussion, conclusion and acknowledgments. Thus, summarization of such documents should be aware of such sectional structure. An intuitive way is to pick a few sentences from each of the sections to be a part of the summary. But how do we decide how many sentences to pick from each section? Also, which sentences to pick? Can we rewrite sentences so as to obtain a concise abstractive summary? We investigate answers to these questions in this paper.

Multiple survey papers have provided a detailed overview of the automatic text summarization task~\cite{tas2007survey, nenkova2012survey,allahyari2017text}. Most of the practically usable summarization systems are extractive in nature. Also, most summarization studies have focused on summarization of news articles. In this work, we mainly focus on two interesting aspects of text summarization: (1) summarization of \emph{scientific research papers}, and (2) summarization for \emph{laymen}. 

~\citet{cohan2018discourse} propose that section level processing of scientific documents is useful. Further,~\citet{collins2017supervised} conclude that not all sections are equally useful. Also, recent papers have observed that a hierarchical summarization of scientific documents is highly effective where at the first level, an extractive summary of each section is independently generated and at the second level, the sectional output is abstracted into a brief summary~\cite{subramanian2019extractive,erera2019summarization}.~\cite{xiao2019extractive} observe that while summarizing, local context is useful, but global is not. Thus, in our approach at the sectional level, we use extracted information from only within the section text to obtain a section's extractive summary, ignoring the remaining text of the entire paper. 


For the LaySumm task, we observe that abstract is the most relevant section of a scientific paper from a layman perspective. We therefore feed the abstract to a Transformer-based model and generate an abstractive summary for the LaySumm task. For the LongSumm task, we first perform extractive summarization for each section and choose a selected number of sentences from each section into the final summary. 

On blind test corpora of 37 and 22 papers for the LaySumm and LongSumm tasks, our proposed system leads to a ROUGE R1 of 45.94 and 49.46 respectively. These results helped us bag the top positions on the leaderboards for the two tasks. 

\section{Related Work}
In this section, we discuss related areas including text summarization and style transfer. 

\subsection{Automatic Text Summarization}
Text summarization focuses on summarizing a given document and obtaining its key information bits. There are two types of text summarization methods: Extractive Summarization and Abstractive Summarization. 

\subsubsection{Extractive Summarization}
Extractive Summarization deals with extracting pieces of text directly from the input document. Extractive Summarization can also be seen as a text classification task where we try to predict whether a given sentence will be part of the summary or not~\cite{liu2019fine}. Most papers in this area focus on the summarization of news articles. But several others focus on specific domains like summarization of medical documents, legal documents, scientific documents, etc. Summarization can also be performed in a query-sensitive manner or a user-centric manner. Sentence-scoring methods include graph-based methods like LexRank~\cite{erkan2004lexrank} or TextRank~\cite{mihalcea2004textrank}, machine learning or deep learning techniques and position-based methods. Recently, various deep learning architectures such as HIBERT~\cite{zhang2019hibert}, BERTSUM~\cite{liu2019text}, SummaRuNNer~\cite{nallapati2016summarunner}, CSTI~\cite{singh2018unity} and Hybrid MemNet~\cite{singh2017hybrid} have been proposed for extractive summarization.

\subsubsection{Abstractive Summarization}

In abstractive summarization, the model tries to generate the summary instead of extracting sentences or keywords. As compared to extractive summarization, this is more challenging and requires strong language modeling schemes to achieve good results. Traditionally, abstractive summarization techniques have focused on generating short text such as headlines or titles. But more recently, there have been efforts on generation of longer summaries. Older methods have depended on tree transduction rules~\cite{cohn2008sentence} and quasi-synchronous grammar approaches~\cite{woodsend2011learning} for effective abstractive summarization. Recently, neural summarization approaches have been found to be more effective. Effective neural representative language models are very important for text generation tasks. With the recent breakthrough of Transformer-based~\cite{vaswani2017attention} architectures like BERT~\cite{devlin2018bert}, T5~\cite{raffel2019exploring} and BART~\cite{lewis2019bart}, utilizing these types of models is crucial for obtaining good textual representations on the target side for neural abstractive summarization.

\subsection{Text Style Transfer}

Neural text style transfer is yet another related area of work where the document in style A is converted to style B without any loss of content or semantics~\cite{syed2020adapting,vadapalli2018science}. This work leverages Transformer encoder-decoder models. The text-encoder is used to obtain robust latent representations while the decoder generates text with a particular target style. 

\begin{figure}
    \centering
    \includegraphics[width=0.95\columnwidth]{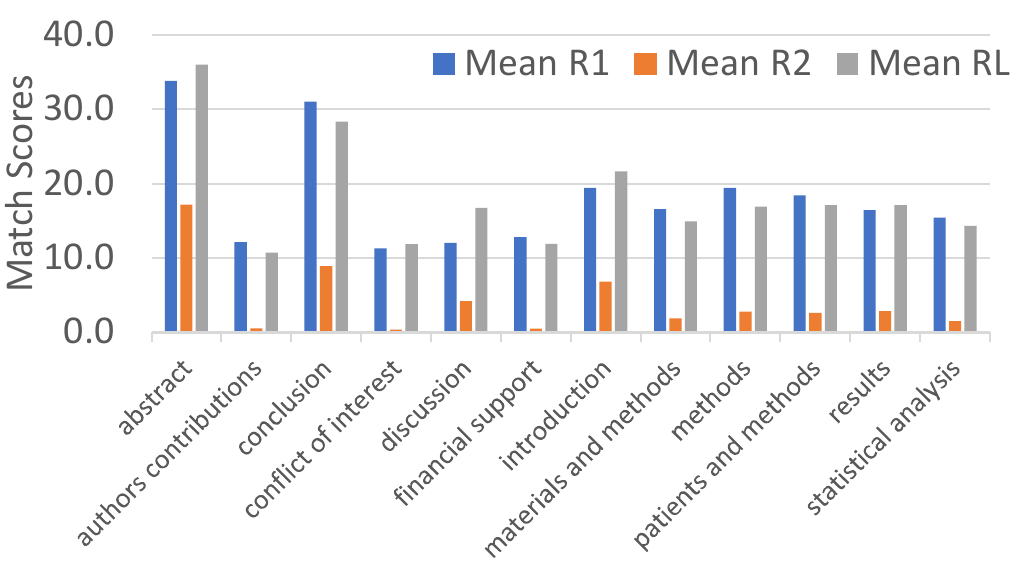}
    \caption{ROUGE-1, ROUGE-2 and ROUGE-L overlaps between paper sections and LaySumm summary}
    \label{fig:LaySummDist}
\end{figure}
\begin{figure}
    \centering
    \includegraphics[width=0.95\columnwidth]{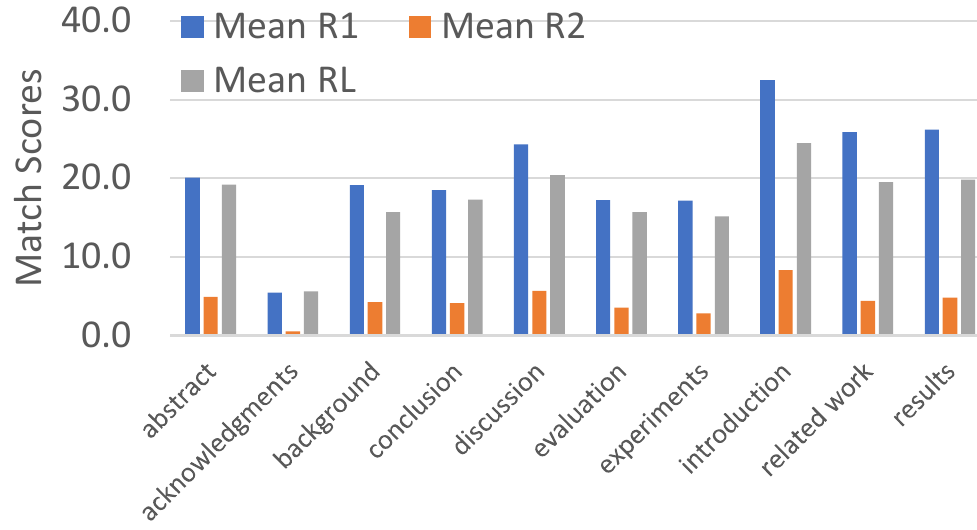}
    \caption{ROUGE-1, ROUGE-2 and ROUGE-L overlaps between paper sections and LongSumm summary}
    \label{fig:LongSummGoldDist}
\end{figure}

\section{Datasets}
We first describe the datasets which were provided by the organizers of the `Workshop on Scholarly Document Processing @EMNLP 2020'\footnote{\url{https://ornlcda.github.io/SDProc/sharedtasks.html}}. 
\subsection{LaySumm Dataset}
A dataset of 572 research papers and corresponding gold standard lay-summaries were available for training, 84 tokens being the average length of a summary. A set of 37 research papers were provided as the blind test data. The LaySumm dataset comprises of full-text papers with lay summaries, in a variety of domains (epilepsy, archeology, and materials engineering), and from a number of journals. Elsevier made available a collection of lay summaries from a multidisciplinary collection of journals, as well as their abstracts and full-texts. For a small sample dataset, look at LaySumm's official GitHub repository\footnote{\url{https://github.com/WING-NUS/scisumm-corpus/blob/master/README_Laysumm.md}}.

\subsection{LongSumm Dataset}

The corpus for this task includes a training set that consists of 1705 extractive summaries, and 531 abstractive summaries of scientific papers in the domains of Natural Language Processing and Machine Learning. The extractive summaries are based on video talks from associated conferences~\cite{lev2019talksumm} while the abstractive summaries are blog posts created by NLP and ML researchers. The average gold summary length was 767 tokens. The research papers were parsed using the science-parse\footnote{\url{https://github.com/allenai/science-parse}} library. A collection of pdfs of 22 research papers served as the blind test set. The LongSumm train and test datasets are publicly accessible on LongSumm's official GitHub repository\footnote{\url{https://github.com/guyfe/LongSumm}}.

\section{System Overview}
In this section, we present an overview of the proposed systems for the LaySumm and LongSumm tasks.

\subsection{System Overview for LaySumm}
We observed that the LaySumm summaries in the train set were highly abstractive in nature with a length limit of 150 words. In Fig.~\ref{fig:LaySummDist}, we analyze how information from each paper section contributes to the final lay-summary by evaluating the ROUGE overlap between a paper section and the available gold summary. This analysis is performed for the entire dataset. Fig.~\ref{fig:LaySummDist} shows that the `abstract' was the most significant section followed by the `conclusion'. Moreover, a relatively high ROUGE-L overlap indicates some degree of verbatim copying from the abstract onto the lay-summary. In addition to providing a high ROUGE overlap, the conclusion section was relatively shorter in length. This indicates that the conclusion section contains a great degree of useful information in a more condensed fashion.

Note that we picked the paper sections directly from the paper text without performing any elaborate conflation on section headers. Conflation in general should not hurt the performance of our models since a particular paper will contain only one form of the section heading, e.g., it will contain either ``materials and methods'' or ``methods''. However, we plan to explore deeper section-wise analysis using improved conflation as part of future work.

We leveraged pretrained Transformer models for conditional generation given a set of individual sections. Our results indicate that using abstract as the only sequence for conditional generation is a better choice as compared to utilizing more sections. Therefore, the problem at hand is one of capturing salient information as one would expect from a summarization task, with the additional flavor of text style transfer.

\begin{figure}[!bth]
    \centering
    \includegraphics[width=0.95\columnwidth]{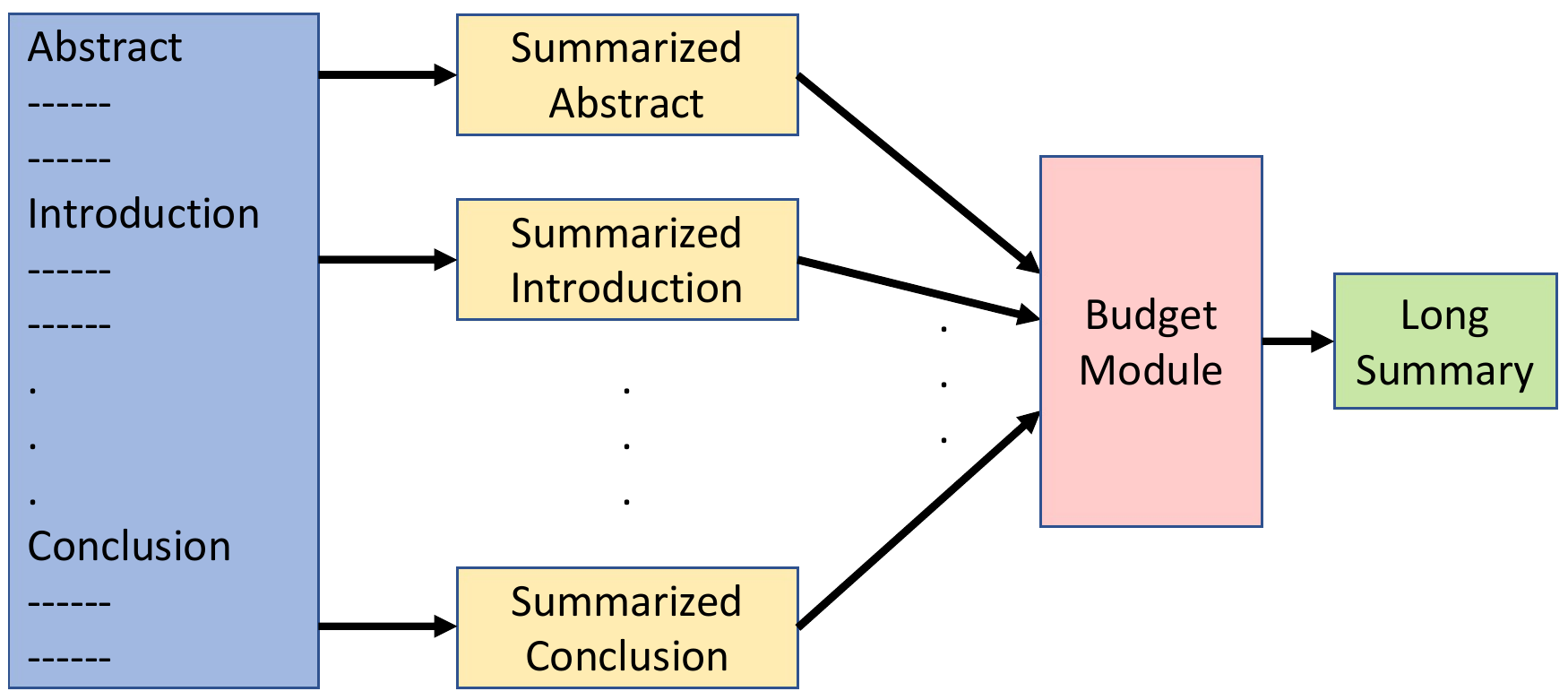}
    \caption{System Architecture for LongSumm}
    \label{fig:arch}
\end{figure}

\subsection{System Overview for LongSumm}
We performed a similar section-contribution evaluation and considered section headings which appeared in at least 5\% of all papers in our training set for the LongSumm task. Fig.~\ref{fig:LongSummGoldDist} shows that the `introduction' is the most important section when it comes to creating summaries followed by `related work' and `results'. 

For our summary generation architecture, we considered one section at a time without the global context. As discussed earlier, this was guided by existing scientific evidence from~\cite{xiao2019extractive} which showed that not considering the global context and focusing purely on the section at hand is marginally better than doing otherwise. Based on section-contribution evaluations, we constructed a budget module to calculate how much weight to assign to a section for the purpose of combining section summaries into the final long-summary. Fig.~\ref{fig:arch} illustrates the broad architecture of our proposed system.

For summarizing each individual section, we used SummaRuNNer~\cite{nallapati2016summarunner}, a simple neural extractive summarizer. We pre-trained SummaRuNNer on the PubMed~\cite{cohan2018discourse} dataset to generate paper abstracts from various paper sections. We show results using variations in our budget module, setting various cutoff thresholds for ROUGE-1 overlap in order for a section to be considered for the summary, i.e., we ignore a section if overlap is less than the threshold value. The system performance indicates that even with a fairly simple neural summarizer at the base, our architecture is capable of achieving superior results on a blind test dataset.

\begin{table*}
    \centering
    \begin{tabular}{|l|c|c|c|c|c|c|}
    \hline
Method&R-1 F1&R-2 F1&R-L F1&R-1 recall&R-2 recall&R-L recall\\
\hline
\hline
Lead-150 baseline&40.85&17.40&25.01&\textbf{54.77}&\textbf{22.96}&\textbf{33.34}\\
\hline
(abs)+SummaRuNNer&39.89&16.30&24.44&51.73&20.89&31.62\\
\hline
(abs)+T5-base&40.74&15.29&24.13&40.32&14.93&23.74\\
\hline
(abs+conc)+T5-base&40.99&15.21&23.72&40.44&14.82&23.28\\
\hline
(abs+conc+intro)+T5-base&40.94&15.32&23.49&40.36&14.92&23.06\\
\hline
(SummaRuNNer)+BART$_L$&40.87&14.72&24.31&43.36&15.46&25.77\\
\hline
(small abs)+BART$_L$&44.81&18.76&26.71&47.61&19.78&28.31\\
\hline
(abs+conc)+BART$_L$&45.45&\textbf{19.22}&27.24&49.56&20.83&29.61\\
\hline
(abs+conc+intro)+BART$_L$&45.69&19.07&27.17&50.17&20.90&29.78\\
\hline
(abs+conc+intro+methods)+BART$_L$&45.61&18.95&27.05&50.50&20.95&29.93\\
\hline
(abs)+BART$_L$&\textbf{45.94}&19.01&\textbf{27.43}&49.11&20.26&29.23\\
\hline
    \end{tabular}
    \caption{LaySumm Results (Best results are highlighted in bold)}
    \label{tab:laySummResults}
\end{table*}

\begin{table*}
    \centering
    \begin{tabular}{|l|c|c|c|c|c|c|}
    \hline
Method&R-1 F1&R-2 F1&R-L F1&R-1 recall&R-2 recall&R-L recall\\
\hline
\hline
Section cutoff at R-1=10.0&47.18&14.10&18.37&43.15&12.93&16.80\\
\hline
Section cutoff at R-1=17.5&49.20&16.49&21.03&44.67&15.00&19.07\\
\hline
Section cutoff at R-1=20.0&48.93&16.57&21.07&\textbf{44.99}&\textbf{15.23}&\textbf{19.36}\\
\hline
Section cutoff at R-1=20.0 + Post-Proc&\textbf{49.46}&\textbf{16.86}&\textbf{21.42}&43.87&14.94&18.98\\
\hline
    \end{tabular}
    \caption{LongSumm Results (Best results are highlighted in bold)}
    \label{tab:longSummResults}
\end{table*}

\section{Experimental Settings}

We used Hugging Face's\footnote{\url{https://huggingface.co/}} implementation of T5 and BART. We experimented with various length settings for training and generation. We found that minimum and maximum sequence lengths of 120 and 140 respectively for generation gave us the best results. We used Adam optimizer with an initial learning rate of 5e-5 with learn rate scheduling based on ROUGE-1 values calculated on the validation split. In the hyper-parameter tuning phase, a repetition penalty of 1.8 while generating lay summaries provided the most optimal results.

We used hpzhao's implementation of SummaRuNNer\footnote{\url{https://github.com/hpzhao/SummaRuNNer}} with default hyper-parameter values. The `topk' parameter was dynamically adjusted to set the summary length of each section based on the section-specific budget. 

\section{Results}
Our system (with the team name: Summaformers) ranks first and third for the LongSumm\footnote{\url{https://aieval.draco.res.ibm.com/challenge/39/leaderboard/39}} and LaySumm\footnote{\url{https://competitions.codalab.org/competitions/25516}} tasks respectively. Further details on these tasks can be found in the shared tasks overview paper~\cite{chandrasekaran2020sdp}. In this section, we present detailed results.  

\subsection{Lay Summary Generation}
We experimented with the BART-large-CNN model which is pre-trained on the CNN/Dailymail summarization dataset and with T5-base in summarization mode. We fine-tuned the conditional generation architectures of these models using the available LaySumm train corpus of 572 documents which we split into training and validation splits in a 4:1 ratio. Our initial results proved the superiority of BART-large-CNN (BART$_L$) over T5. We experimented with various generative sources such as abstract only, abstract + conclusion, abstract + conclusion + introduction, abstract + conclusion + introduction + methods. Furthermore, owing to the structure of the abstract itself, we considered the first, second and final paragraphs of the abstract (also referred to as ``small abs'') as the source. Our results (Table~\ref{tab:laySummResults}) show that using the complete abstract as input to BART$_L$ is the best performing setting for LaySumm. Since the papers in the dataset were published in various scientific journals, the original abstracts contain highly domain-specific technical jargon. The BART$_L$ model captures the salient points from the abstract in a short 150-word budget while transferring the text style from scientific to a layman style. 

After hyper-parameter tuning on the generation end, we achieved a ROUGE-1 score of 45.94 on the blind-test corpus. Our generated summaries are coherent in addition to being highly abstractive in nature.

For comparison, we also present results for a na\"ive ``Lead-150'' baseline which outputs the first 150 tokens of the abstract as the summary. As shown in Table~\ref{tab:laySummResults}, surprisingly, this simple baseline leads to impressive results especially on recall metrics. Running SummaRuNNer on the abstract leads to results which are worse than the Lead-150 baseline.  

\subsection{Long Summary Generation}
We used the SummaRuNNer~\cite{nallapati2016summarunner} neural extractive summarization system as our base section summarizer. We pretrained this on the training set of the publicly available PubMed dataset (using GloVe~\cite{pennington2014glove} 6B 100D word embeddings) to generate the paper abstract as closely as possible from any given section. This grounds the network in a setting where it can easily capture salient points. We plan to explore pretraining with other datasets as part of future work.

This was further finetuned using the LongSumm train set as follows. The given LongSumm training dataset was divided into train and validation splits in a 9:1 ratio. We used the same previous settings to finetune on documents in the LongSumm train split. Now, the pretrained SummaRuNNer model was conditioned to extract sentences which maximize the ROUGE-1 overlap with the provided gold standard long summaries.

Finally, based on our budget module, we assign a weight to each available section and generate section summaries of computed lengths which are further concatenated to generate the final summary. We experiment with various settings in the weight assignment based on specified overlap cutoffs in the budget module as shown in Table~\ref{tab:longSummResults}. The best performing setting corresponds to selecting sections whose ROUGE-1 overlap with the long summary is greater than 20.0. Intuitively, this prunes out irrelevant sections such as `abbreviations' and `acknowledgements'. The remaining sections were assigned weights based on the ROUGE-1 overlap with the provided long summary. The generated long summaries are extractive and capture the most salient pieces of information from the given research papers. The results improve slightly when we perform ad hoc post-processing using heuristics like removing paper citations within brackets, removing non-English Unicode characters and mathematical notation symbols.

\subsection{Case Studies}
In the following, we present two cases of lay summaries generated by our system. As we can see, the generated summaries are highly abstractive and coherent. They also capture the important aspects of the paper.

For the article at this URL\footnote{\url{https://doi.org/10.1016/j.engappai.2020.103495}}, the generated lay summary was as follows: `This paper proposes a novel approach to support the transformation of bioinformatics data into Linked Open Data (LOD). It defines competency questions that drive not only the definition of transformation rules, but also the data transformation and exploration afterwards. The paper also presents a support toolset and describes the successful application of the proposed approach in the functional genomics domain. According to this approach, a set of competency criteria drive the transformation process. This paper presents a framework for the development of an open data management system that can be easily adapted to different data types.'

For the article at this URL\footnote{\url{https://doi.org/10.1016/j.engappai.2019.103467}}, the generated lay summary was as follows: `To foster interaction, autonomous robots need to understand the environment in which they operate. One of the main challenges is semantic segmentation, together with the recognition of important objects, which can aid robots during exploration, as well as when planning new actions and interacting with the environment. In this study, we extend a multi-view semantic segmentations system based on 3D Entangled Forests (3DEF) by integrating and refining two object detectors, Mask R-CNN and You Only Look Once (YOLO), with Bayesian fusion and iterated graph cuts. The new system takes the best of its components, successfully exploiting both 2D and 3D data.'

Finally, the following lay summary was generated by our model for this very paper: `In this paper, we develop a novel system for summarizing scientific research papers from multiple domains. We differentiate between two types of summaries, namely, (a) LaySumm : a very short summary that captures the essence of the research paper in layman terms restricting overtly specific technical jargon and (b) LongSumm a much longer detailed summary aimed at providing specific insights into various ideas touched upon in the paper. While leveraging latest Transformer-based models, our systems are simple, intuitive and based on how specific paper sections contribute to human summaries of the two types described above.'

\section{Conclusions}
In this paper, we studied two scientific document summarization tasks: LaySumm and LongSumm. We experimented with popular text neural models in a section-aware manner. Our results indicate that modeling of the document structure with strong focus on which parts of a research paper to attend to while composing a summary gives a significant boost to the quality of the resultant output. On blind test corpora, our system ranks first and third for the LongSumm and LaySumm tasks respectively.
\bibliographystyle{acl_natbib}
\bibliography{emnlp2020}

\end{document}